# Feature Interactions in XGBoost

Kshitij Goyal, Sebastijan Dumancic, Hendrik Blockeel

KU Leuven, Belgium

**Abstract.** In this paper, We investigate how feature interactions can be identified to be used as constraints in gradient boosting tree models through XGBoost's implementation. Our results show that accurate identification of these constraints can help improve the performance of baseline XGBoost model significantly. Further, improvement in the model structure can also lead to better interpretability.

## 1 Introduction

In statistics, two variables $X_1$ and $X_2$ are said to interact when the effect of $X_1$ on a target variable $Y$ is not constant, but may depend on the value of $X_2$. "Effect" here refers to the increase or decrease of $Y$ that is caused by a change in $X_1$ (while keeping $X_2$ constant). Interaction does not occur in linear models of the form

$$Y = aX_1 + bX_2$$

an unit increase of $X_1$ by 1 will always increase $Y$ by $a$ (assuming $X_2$ is kept constant). On the other hand, a model of the form

$$Y = aX_1 + bX_2 + cX_1X_2$$

does have interaction: rewriting it as

$$Y = (a + cX_2)X_1 + bX_2$$

makes clear that a unit increase of $X_1$ increases $Y$ by $a + cX_2$, hence, by a value that depends on $X_2$. In statistics, a common way to detect interaction is to train models with and without product terms, and check whether the version with a product term significantly outperforms the one without.

In the context of decision trees, $X_1$ and $X_2$ do not interact if the following holds: whenever a test of $X_2$ is used, either $X_2$ does not occur in any of the branches, or if it does, it must be tested in exactly the same way in all branches. Indeed, if we would have a test $X_1$<a where the left branch contains $X_2$<b and the right one does not, this means that the target depends on $X_2$ if $X_1$<a, but does not depend on it if $X_1$>a; that is, the effect of $X_2$ on $Y$ clearly depends on the value of $X_1$. Obviously, having identical sub-trees in the left and right branch is relatively rare. So, typically, a tree will either not use a variable, or use it in a way that allows for interaction with other variables.

Ensembles of decision trees, however, make it much easier to use variables and yet disallow interaction. If two variables $X_1$ and $X_2$ are used in different trees, but never in the same tree, then they do not interact: $X_1$ can only affect the prediction through the trees it occurs in, and since $X_2$ does not occur in these trees, the value of $X_2$ cannot possibly affect the effect of $X_1$, and vice versa. So whereas trees are biased towards interaction, ensembles of trees introduce a natural way of including both variables that do interact (by having them in the same tree) and variables whose effects do not interact (by having them in different trees).

The popular XGBoost system exploits this fact. It allows the user to partition the variables into subsets of variables among which there may be interaction, referred to as 'feature interaction constraints'[1]. Given a split on feature $f$, only the features belonging to the same subset as $f$ will be considered for next split in the tree. By exploiting this constraint, the system reduces its search space. This may lead to higher-quality, more efficient models with less noise in predictions, better generalization and high interpret-ability. If a dataset has a high number of features, identifying these constraints can help to make the modelling more directed, which makes the interpretations of the model more accurate. There are two main aspects of interpretability, feature importance and feature interactions. In gradient boosting, feature importance can be measured using metrics like gain, weight and cover or by calculating a relevance measure [1]. Also, interpretability in terms of feature interaction has been studied in [2]. However, There aren't a lot of interpretability methods which directly address feature interactions in gradient boosting trees. Knowing these interaction constraints in advance can fill that gap. Having identified important features using the existing methods, these partitions can be used to interpret the interactions as the tree ensembles are restricted to comply with the constraints.

However, the user has to provide these constraints. An interesting question is then: can we automate the creation of these constraints by first analysing the data, looking for interacting variables, and then providing this information to XGBoost? In this paper, we provide some initial results that suggest a positive answer.

XGBoost models are state of the art and outperform many other machine learning methods[2]. A significant improvement on these models could be substantial. There has already been a lot of work on feature selection methods which use the idea of feature interaction [[4], [6], [5], [10]], but none on specifically identifying these constraints. In our work, we aim to develop methods to identify these constraints and study their effects on model performance and efficiency. Identifying these kind of feature interactions can be useful in other machine learning algorithms, specifically decision tree based algorithms. These constraints can be incorporated as a part of existing ensemble solutions (XGBoost, LightGBM etc.) to create a more automated solution with better interpretations of the modelled features.

---

[1] https://xgboost.readthedocs.io/en/latest/tutorials/feature_interaction_constraint.html

[2] *XGBoost - ML winning solutions (incomplete list)*

## 2  Approach

Feature Interaction has mainly been studied in the context of feature selection in the current literature. There are two main types of methods for feature selection, filter methods and wrapper methods[7]. Filter methods use the general features of the dataset like correlation, on the other hand, wrapper methods use simpler machine learning algorithms to identify important features. Filter methods are faster and slightly less accurate compared to wrapper methods which are computationally more expensive[3].

In our work to identify feature interaction, we explored both types of methods. Starting with a filter method based on mutual information, we realized that this method is not able to capture the higher level interaction of decision trees. We worked on an iterative selection approach based on mutual information inspired by the algorithm described in [4], with underwhelming results. We are not presenting this method in the paper. However, we present a wrapper algorithm which uses linear or logistic regression models to identify feature interaction.

### 2.1  Wrapper Algorithm

This algorithm uses an iterative approach to get to the splits of the full feature set. The splits identified by this method are mutually exclusive.

- Start with a full set of features F
- Fit a linear/logistic regression model for each of the feature depending upon the nature of the target, taking one feature at a time. Select the feature that performs the best. Add it to the selected feature list and remove from the full feature list.
- From the remaining features, select the feature where the model with interaction performs better than the model without interaction and the model with interaction is best compared to other remaining features. Model with interaction has all the possible pairwise interactions. Add it to the selected feature list and remove from the full feature list.
- If no new feature can be added to the selected feature list, add the selected feature list to the final interaction list and repeat the process from step 2 on the remaining of the original feature list, until there aren't any features left.

```
Result: Interaction splits of the full Feature list
Features = full feature set;
InteractionSplits = [];
Subset = [];
while len(Features) != 0 do
    PotentialCandidates = [];
    for f_i in Features do
        Model = Logistic/Regression model with Subset∪f_i;
        ModelI = Logistic/Regression model with Subset∪f_i with
         interactions;
        if performance(Model_I) > performance(Model) then
         |  add f_i to the PotentialCandidates;
        end
    end
    if len(PotentialCandidates) > 0 then
        f_b = Select the best performing feature from PotentialCandidates;
        Add f_b to Subset;
        Remove f_b from the Features;
    else
        Add subset to InteractionSplits;
        Subset = [];
    end
end
```
**Algorithm 1:** Wrapper Algorithm for Feature Interaction

## 3 Experiments

Using the approach for finding the split, we have created multiple gradient boosting models using XGBoost for a number of data-sets for both classification and regression problems. Gradient boosting works by combining multiple weak learners into a strong learner in an iterative fashion. At every step the residual is modelled using a decision tree until the improvement is significant[8].

### 3.1 Modelling

We have created following variations of Gradient Boosting Tree models for comparison:

– **Baseline:** The original XGBoost model without any feature interaction constraints.
– **Full Interaction Model:** XGBoost Model which uses interaction splits identified by the original target in all the trees in the ensemble.
– **Partial Interaction Model:** XGBoost Model with only a selected number of trees in the beginning of the ensemble having interaction constraints. For each residual tree, new interaction splits are calculated based on individual residual targets. The number of trees with interaction is also a parameter and we have modelled different variations under 30.

We use cross validation to tune parameters: Number of trees, maximum depth and learning rate.

### 3.2 Results

To demonstrate that these feature interaction constraints can actually help make the XGBoost model more accurate, we compared the baseline model (model without any interaction constraints) to the models with randomly generated interaction constraints. We did this for two publicly available data sets and in 20 attempts we could find 4 and 3 constraints which perform better than baseline, for 'cleve' and 'ada_prior' respectively. Results shown in Table 1 indicate the value in utilizing these feature interaction constraints.

| Dataset | Interaction Constraints | Baseline Accuracy | Model Accuracy |
|---|---|---|---|
| cleve | [[3, 10, 1, 9, 2, 8], [6, 7, 4, 11, 0, 5, 12]] | 84.615385 | 85.714286 |
| cleve | [[2, 8, 12, 4, 0, 9], [3, 6, 5, 1, 7, 10, 11]] | 84.615385 | 87.912088 |
| ada_prior | [[12, 11, 8, 10, 1, 2, 13], [6, 7, 3, 5, 4, 9, 0]] | 84.214445 | 84.735666 |
| ada_prior | [[3, 0, 7, 11, 2, 1, 6], [12, 4, 8, 10, 13, 9, 5]] | 84.214445 | 84.437826 |

**Table 1.** Random Feature Interactions

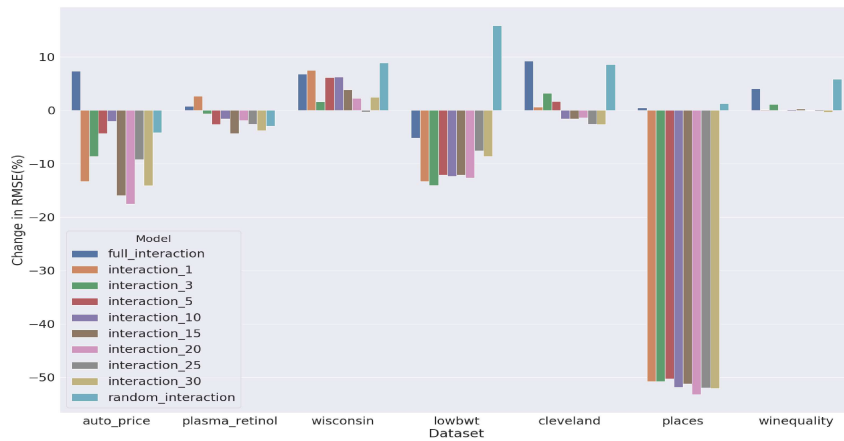

**Fig. 1.** Regression Output Logistic Wrapper - Change from the baseline

Next, we have applied the logistic wrapper on 9 classification data-sets and 7 regression data-sets to identify interaction constraints. These data sets are obtained from openml repository[3]. Data-sets with high number of features (>=10) are selected to have more possibility of higher order interactions. In Figures 1 and 2, x-axis represents the dataset and y -axis represents the percentage change from the baseline. 'full_interaction' represents the model with with a single interaction constraint throughout the ensemble. Models represented as 'interaction_x' are partial interaction models with interaction constraint on the first 'x' trees in the ensemble. Models represented as 'random_interaction' are average over 5 different models with randomly generated interaction constraints.

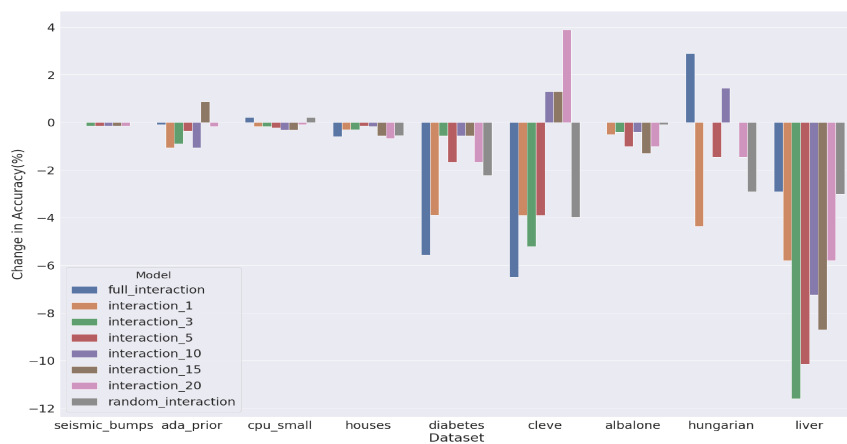

**Fig. 2.** Classification Output Logistic Wrapper - Change from the baseline

- For regression data sets (Fig. 1), we see significant improvement for three of the data sets compared to baseline and random interaction models. Partial interaction models perform much better than the full interaction models in most of the cases. This suggests that residuals have different relationships with the features than the original target, utilizing these relationships is better than having a single constraint for the full ensemble.
- For classification data sets (Fig. 2), our approach under-performs significantly with reduction in accuracy for most of the data sets. This is something that we plan to work on next.

## 4  Conclusion and Future Works

As discussed in the introduction, we identify the value in utilizing the feature interaction constraints in XGBoost. With the correct feature interaction infor-mation, we can get a significant improvement on XGBoost. In our work, we have tried to develop approaches to do that. In the wrapper method we presented, we get significant improvement over the XGBoost models in a number of different data sets for regression problems. In future, we aim to:

- Understand the drawbacks of our method and improve for the classification problems.
- Modify our approach to perform at least as good as the baseline and analyze the efficiency of constrained models.
- Study the change in interpretability of the model compared to baseline.
- Explore approaches like association rules to group interacting features.

---

[3] https://www.openml.org/search?type=data